\documentclass[conference]{IEEEtran}
\IEEEoverridecommandlockouts
\usepackage{cite}
\usepackage{amsmath,amssymb,amsfonts}
\usepackage{algorithmic}
\usepackage{graphicx}
\usepackage{multirow}
\usepackage{textcomp}
\usepackage{xcolor}
\def\BibTeX{{\rm B\kern-.05em{\sc i\kern-.025em b}\kern-.08em
    T\kern-.1667em\lower.7ex\hbox{E}\kern-.125emX}}

\usepackage[textsize=scriptsize, textwidth=100]{todonotes}
\setuptodonotes{color=blue!30}
\newcommand{\newauthor}[3]{\newcommand{#1}[1]{\todo[color=#3]{#2: ##1}}}
\newauthor{\christian}{cg}{lime}
\usepackage{lipsum}  

\title{DADO -- Low-Cost Query Strategies for \\ Deep Active Design Optimization
\thanks{This research has been funded by the Federal Ministry for Economic Affairs and Climate Action (BMWK) within the project "KI-basierte Topologieoptimierung elektrischer Maschinen (KITE)" (19I21034C).}
}

\usepackage{svg}
\usepackage{amsmath}
\usepackage{hyperref}
\usepackage{float}
\usepackage{subcaption}
\usepackage{colortbl}
\usepackage{standalone}
\usepackage{tikz}
\usepackage{pgfplots}
\pgfplotsset{compat=1.18}
\pgfplotsset{paraviewStyle/.style={%
hide axis,scale only 
axis,height=0pt,width=0pt,xmin=0,xmax=1,ymin=0,ymax=1,domain=0:1, 
point meta=f(x),
}}

\author{\IEEEauthorblockN{
Jens Decke\IEEEauthorrefmark{1},
Christian Gruhl\IEEEauthorrefmark{1},
Lukas Rauch\IEEEauthorrefmark{1},
Bernhard Sick\IEEEauthorrefmark{1},
}
\IEEEauthorblockA{\IEEEauthorrefmark{1}Universit\"at Kassel, Germany, Email:  (jens.decke$|$cgruhl$|$lukas.rauch$|$bsick)@uni-kassel.de}
}

\begin{document}

\maketitle

\begin{abstract}
In this experience report, we apply deep active learning to the field of design optimization to reduce the number of computationally expensive numerical simulations.
We are interested in optimizing the design of structural components, where the shape is described by a set of parameters. If we can predict the performance based on these parameters and consider only the promising candidates for simulation, there is an enormous potential for saving computing power.
We present two query strategies for self-optimization to reduce the computational cost in multi-objective design optimization problems. Our proposed methodology provides an intuitive approach that is easy to apply, offers significant improvements over random sampling, and circumvents the need for uncertainty estimation. 
We evaluate our strategies on a large dataset from the domain of fluid dynamics and introduce two new evaluation metrics to determine the model's performance.
Findings from our evaluation highlights the effectiveness of our query strategies in accelerating design optimization.
We believe that the introduced method is easily transferable to other self-optimization problems.
\end{abstract}


\section{Introduction}
High-performance computing (HPC) systems have a high energy demand, which results in significant carbon dioxide emissions contributing notably to climate change~\cite{Gupta2021Chasing}. Numerical simulations, for instance, computational fluid dynamics~(CFD) or finite element analysis~(FEA), widely used in industry and research, often demand days to weeks of computing time on HPC systems, posing a particular concern in this regard. Examples include simulations for weather predictions, structural dynamics, and electrodynamics. Reducing the number of numerical simulations or accelerating them could lead to significant savings in energy consumption and the reduction of carbon dioxide emissions.

Design optimization (DO) aims to determine the optimal shape of components and typically involves many numerical simulations to identify the best design for pre-defined constraints. In recent years, deep learning methods are emerging in the field of DO to accelerate numerical simulations or to improve the overall performance~\cite{Calzolari2021Deep}. Nevertheless, there is still the need for massive annotated training datasets. The annotations are acquired through numerical simulations, which are computationally expensive. To tackle this problem, we propose an approach to reduce the number of computer simulations required in DO processes with deep active learning~(DAL) for regression. We refer to this as deep active design optimization~(DADO).

In DAL, the objective is to train a deep learning model, while actively selecting and annotating the most useful samples from a non-annotated dataset~\cite{Kottke2017Challenges}. The criteria for determining the most valuable sample depends on the specific application area and will be explicitly defined later for our use case. Unlike traditional passive learning approaches, which require a large amount of annotated data, DAL aims to reduce the annotation effort by iteratively selecting the most useful samples for annotation. In DAL, the selection is typically based on query strategies, such as uncertainty or diversity sampling~\cite{Kumar2020Active}. These strategies aim to identify samples that are expected to improve the model's performance the most. The selected samples are then annotated by an oracle, which could be a human expert or a computer simulation. The annotated samples are used to update the deep learning model, incorporating the newly annotated samples into the training process. This iterative cycle of self-selecting samples, annotating samples, and updating the model continues until a pre-defined stopping criterion is achieved. The main advantage of DAL for regression is the potential to achieve high performance with less annotated data compared to traditional supervised learning approaches~\cite{Ren2021A}.

We conduct experiments on a real-world DO use case (cf. Figure~\ref{fig:usecase}) in the problem domain of fluid dynamics and thermodynamics, where flow deflections significantly contribute to efficiency losses in technical systems, such as piping systems for industrial heating and cooling. The objective is to discover a design that both reduces pumping power and ensures sufficient cooling capacity. This is a typical multi-objective and multi-physics optimization problem. Our approach employs DAL to reduce the number of computer simulations required for DO by selecting only the most valuable samples (i.e., those that are expected to yield the best performance gains), rather than accelerating individual simulations. We begin with a small number of randomly drawn annotated samples (i.e., designs) and a large data-pool of non-annotated samples (i.e., design candidates).
The query strategy iteratively selects design candidates to be evaluated by the computer simulation (i.e., expert model) to provide the ground truth annotation. The objective of this approach is to maximize performance with as few requests to the expert model as possible by selecting only those design candidates that are expected to be the most valuable for the model's performance.

In typical DAL scenarios, the primary objective is to attain high predictive performance across the entire dataset.
In DADO, the primary objective is to find a multi-objective optimal solution with as few candidate evaluations as possible.
Since we are only interested in promising candidates, the predictive performance must only be high for these candidates, and it is not necessary to discriminate between mediocre and bad candidates. 
Consequently, our interest lies in a prediction model that exhibits strong performance and generalization within the feature space (i.e., design space) where the optimal solution is likely to reside. Metaphorically, this concept can be linked to a shrouded mountain range, where the peaks of different mountains emerge above a dense layer of fog. Rather than focusing on the entirety of the mountain, we solely concentrate on the elevated summits. One challenge in DAL is that the selection of promising design candidates for annotating can be biased towards certain regions of the design space~\cite{Ren2021A} which results in bad model generalization. In contrast, we deliberately induce a bias by exploiting only the most promising regions in the design space.
Thus, since conventional query strategies are not well-suited to address our primary objective, we have developed two  low-cost query strategies that enable a model training within the relevant design space. They are characterized by their ease of implementation, low computational cost, and high effectiveness in finding promising design candidates. We refer to them as L2-Select and L2-Reject, as they select or reject design candidates based on the L2 norm. The proposed query strategies are also applicable to other self-optimization problems and can be used to guide decision-making. Additionally, we propose two metrics tailored to the DAL regression problem to monitor and evaluate the model's performance at each iteration. Two scenarios with high and low annotation budgets with different DAL experiment parameters are investigated.
\newline 

This experience report presents our proposal to address DO problems using DAL methods. In addition to the publicly available code \footnote{\url{https://git.ies.uni-kassel.de/jdecke/lcs4dado}} developed on an open access dataset~\cite{decke2023dataset} we provide reproducible experiments and the following \textbf{contributions} to the research area.
\begin{itemize}
\item We conduct initial research in applying DAL in the domain of DO as an optimization method to efficiently discover promising design candidates, therefore reducing the number of numerical simulations.
\item We propose two novel low-cost query strategies for multi-objective DADO. Additionally, we introduce two metrics to evaluate the model's performance.
\item We make the first steps towards a deep generative active learning-based model for DO. The report also presents and discusses the challenges we encountered during this process.
\end{itemize}

The remainder of this article is structured as follows. Section~\ref{sec:related_work} briefly overviews related work, focusing specifically on deep learning in design optimization and active learning. Section~\ref{sec:design-optim} delves into the considered problem domain, providing a concise discussion on design optimization and focusing on the domain of fluid dynamics. Furthermore, we introduce our dataset, outlining its relevance to our research. Moving on to Section~\ref{sec:method}, we present our methodology in detail, describing how we trained a deep neural network and highlighting the query strategies employed. Section~\ref{sec:experimental} is dedicated to the experimental setup and its results, where we compare our newly developed query strategies against random strategies, providing insightful analyses and statistical observations. In Section~\ref{sec:outlook}, we present an idea to extend the described method to include a variational autoencoder~(VAE) for future research. Finally, the article is concluded in Section~\ref{sec:conclusion}.

\section{Related Work} \label{sec:related_work}

The optimization of design is a fundamental problem in engineering that has been extensively investigated over several decades~\cite{Martins2013MultidisciplinaryDO}. Recently, there is a growing interest in employing machine learning methods to study DO problems. This interest is spurred by two factors: first, the emergence of new additive manufacturing techniques, which enable the production of free-form designs~\cite{Nguyen2017Topology}; and second, the availability of computing power that allows the resolution of complex and relevant industrial problems~\cite{Martins2013MultidisciplinaryDO}. For example, a current study shows the possibilities of combining DO and additive manufacturing of electromagnets~\cite{Bechari2023From}.
Nie et al.~\cite{Nie2021TopologyGAN} proposed \textit{TopologyGAN} in 2021. It is used to optimize the stress and strain in a simple truss structure by comparing it with a baseline conditional generative adversarial network. The authors generate a dataset comprising already optimized truss structures, which were dependent on the size and direction of the load. The model's generalization capability was evaluated by applying unknown load boundary conditions. Although \textit{TopologyGAN} did not perform optimization, it was able to identify an optimal truss structure for changed boundary conditions.
The authors of~\cite{ripken2022learning} employed a graph neural network; with knowledge of the boundary conditions, they aim to generalize to previously unobserved or superimposed numerical meshes.
A study from 2022 investigates if \textit{anomaly detection} algorithms can be used to solve DO problems~\cite{decke2022ndnet}. A significant problem is the tradeoff between exploration and exploitation. The key finding is that \textit{anomaly detection} can be used to explore the design space. Still, there is a great difficulty in exploitation because anomaly detection algorithms would consider a design candidate as already detected whose target value is only slightly better than an already known one. The methodology in this work seeks to focus on exploitation without compromising exploration.
Genetic algorithms (GA) such as the Non-dominated Sorting Genetic Algorithm 2 are well-established methods for solving DO problems; however, their convergence speed is rather slow~\cite{Deb2002a, Duan2009Comparsion}. 
In 2022, Parekh et al. developed a generative model for electrical machines with multiple topologies by using VAE in conjunction with a prediction network~\cite{parekh2022Variational}. They concatenated the design parameter spaces of two distinct machine topologies and trained a latent representation that was highly effective in reconstructing the input. The latent dimension employed was defined to be greater than the design parameter space of the more complex machine topology in the latent space. Consequently, the latent representation did not compress any information of the input, and we hypothesize that the network learned the identity of the input designs only. The prediction network extended the capabilities of the VAE to enable it to predict objective values in a supervised manner. The dataset of both machines used in their study included $28,278$ designs, which is a considerable amount of data. In real-world scenarios, DO problems do not typically provide such a large dataset. So our approach aims to use a significantly smaller number of design candidate with the help of DAL without compromising the model's prediction performance. Unfortunately, it was not possible to reproduce and extend their ideas because the code and data were not publicly available.

To the best of our knowledge, DAL was not yet directly applied in DO. Nevertheless, Deng et al. introduce a comparable approach called \textit{Self-directed Online Learning Optimization} for topology optimization in 2022~\cite{Deng_2022self}. This approach integrates neural networks (NN) with numerical simulation data. The NN learns and substitutes the target as a function of design variables. At the same time, a small subset of training data is generated dynamically based on the NN prediction of the optimum. The NN fits the new training data and provides a better prediction in the region of interest until convergence. New design candidates selected for numerical evaluation are generated by adding a disturbance to the best design of the previous iteration, similar to mutation and crossover of GA. The main difference between the work of Deng et al. and this article is how the query strategy performs. We focus on low-cost query strategies, while they added disturbance to their design parameters. Furthermore, we have a vast dataset available to conduct our experiments offline. A request to the computer simulation can be answered instantaneously by drawing a design from the data-pool.

\section{Use Case}
\label{sec:design-optim}
The DO methodology developed in this work is based on a use case from the field of fluid dynamics and thermodynamics, but can also be applied to other problems and domains such as aerospace engineering, automotive industry, civil engineering, architecture, and manufacturing. In aerospace engineering, DO is used to improve the performance and efficiency of aircraft components, such as wings, fuselage, and engines. In the automotive industry, DO is employed to enhance the performance and safety of vehicles, such as improving aerodynamics, reducing emissions, and increasing efficiency of electromagnets~\cite{Lucchini2022Topology}. In civil engineering, DO is applied to optimize the design of structures such as bridges, buildings, and dams, in terms of strength, stability, and cost. In architecture, DO is used to improve building performance regarding energy efficiency, natural light, and structural integrity. In manufacturing, DO is employed to optimize the design of products, such as reducing material waste and improving production efficiency.

\begin{figure}[ht]
	\centering
    \includegraphics[width=.25\textwidth]{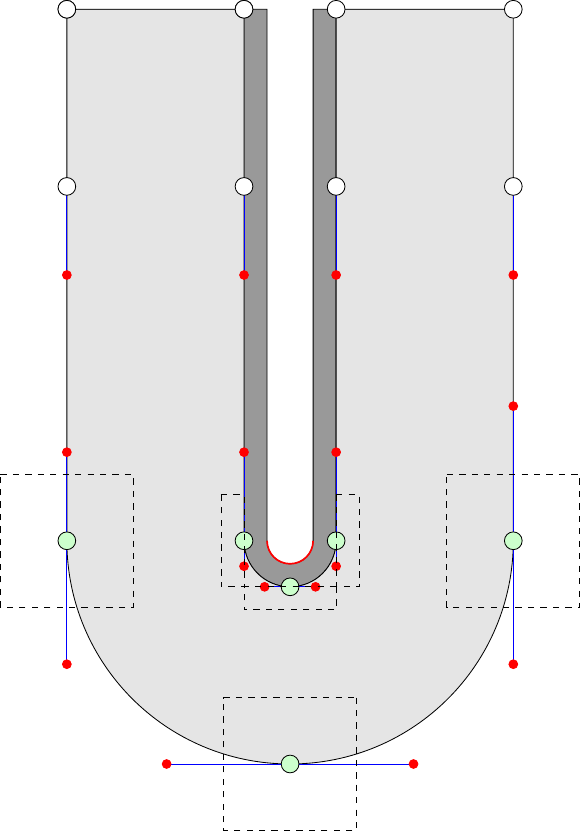}
	\caption{Parameterized geometry with 6 boundary points (defined by two design parameters each) in green and 16 curve parameters in red (adapted from~\cite{decke2023dataset}). The objective of this use-case is to find a U-Bend design with minimal pressure loss and high cooling power.}
    \label{fig:usecase}
\end{figure}

Our use case is a U-Bend flow channel. They can be found in various technical systems and applications, particularly those involving fluid transport or heat transfer. They are commonly employed in heat exchangers, such as condensers and evaporators, where they facilitate the transfer of heat between a fluid and its surroundings. U-bend flow channels can also be utilized in piping systems, refrigeration systems, air conditioning systems, and hydraulic systems to redirect or control the flow of fluids. The parameterization of the U-Bend is depicted in Figure~\ref{fig:usecase}. It is described with 28 design parameters and two target values.

\begin{figure}
    \centering
    \begin{tikzpicture}
        \begin{axis}[
            colorbar horizontal,
            colormap/jet,
            colorbar style={
            title={Pressure [Pa]}, 
            width=0.35\textwidth, 
            height=6, 
            xtick={-80,-50,-20,10,40},
            point meta min=-80, 
            point meta max=40 
            },
            paraviewStyle,
            ]
        \end{axis}
    \end{tikzpicture}
    \includegraphics[width=.25\textwidth]{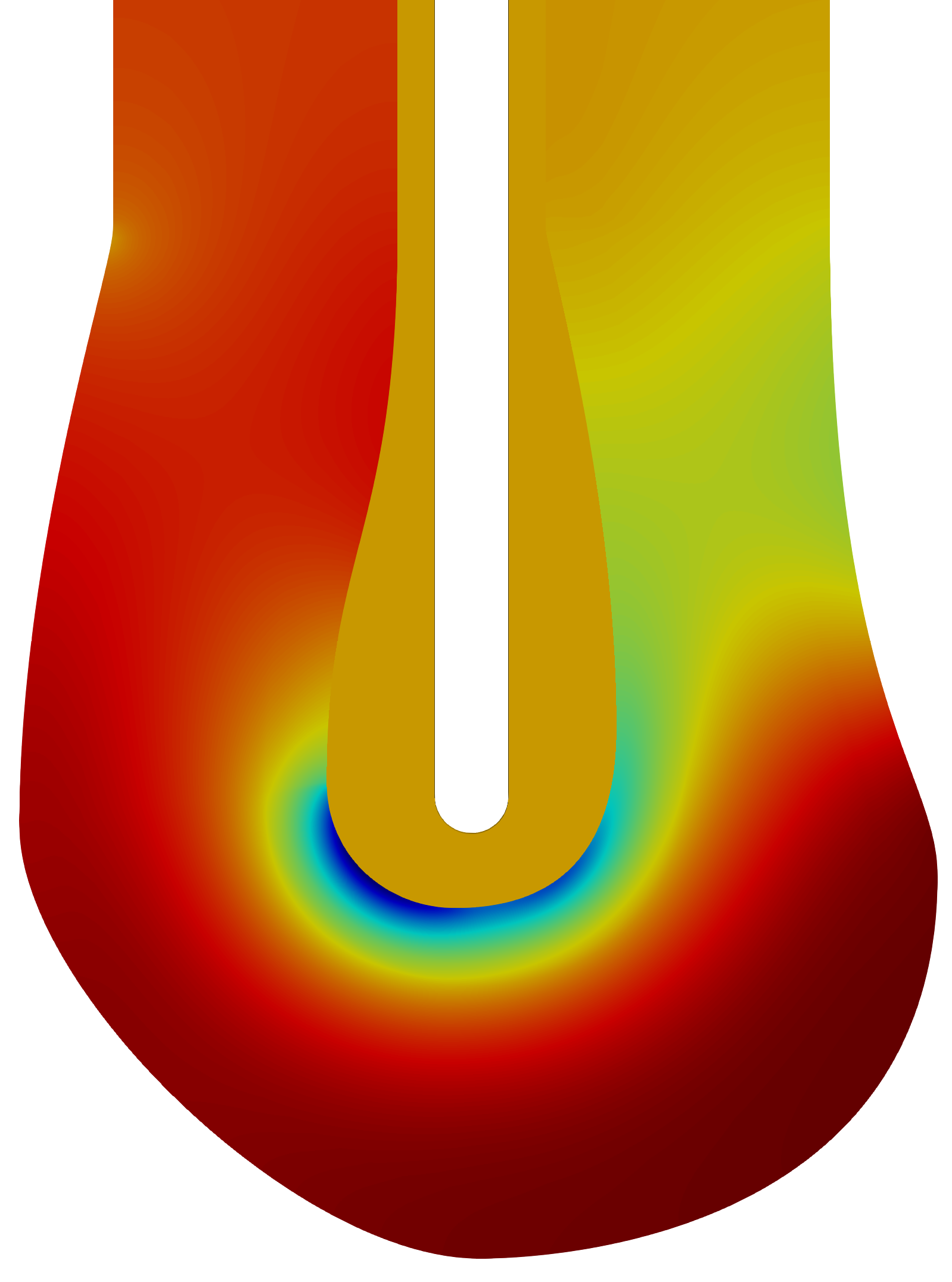}
    \caption{Example of a certain design candidate solved to compute the pressure using a computationally expensive numerical simulation. The scalar target value pressure loss is derived from this result representation.}
    \label{fig:usecaseobjective}
\end{figure}

The parameterized geometry utilizes six boundary points, illustrated in green, with each boundary point offering two design parameters that are allowed to vary within their respective dashed bounding boxes. Additionally, we incorporate 16 curve parameters to connect these boundary points. In Figure~\ref{fig:usecaseobjective}, we present exemplary the pressure distribution of a particular design candidate, obtained through numerical simulation using the expert model. In a subsequent post-processing analysis, the pressure loss is computed based on this simulated solution.

The design parameters determine the shape of the flow deflection, while the target values represent the pressure loss in [$Pa$] and the cooling capacity, which is quantified as the squared temperature difference between the heating surface and the cooling medium in [$K^2m^2$]. A small temperature difference corresponds to a high cooling capacity. The dataset comprises three distinct data formats for each design. However, for the purpose of this study, our focus lies solely on the parameter representation of the designs. This particular representation is chosen due to its streamlined and efficient nature, making it ideally suited for our methodology. The data is freely available and can be found in~\cite{decke2023dataset}, providing additional information on this specific use case and the numerical investigations to obtain the data.

\section{Methodology}
\label{sec:method}
\subsection{DAL Process}
\begin{figure}[ht!]
    \centering
    \includegraphics[width=.48\textwidth]{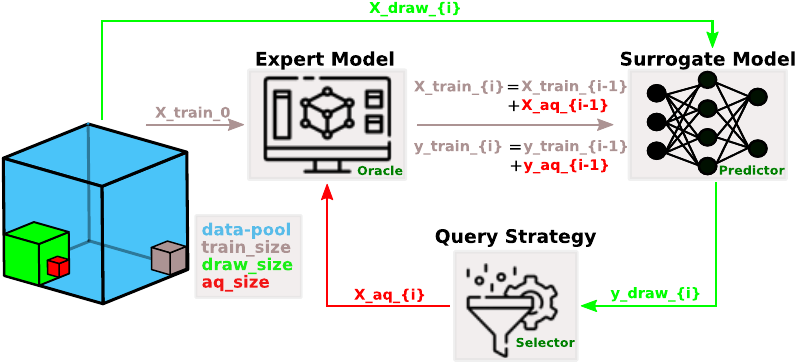}
    \caption{The DAL process: The \textbf{Expert Model} is initially trained on $X\_train\_0$ and executes expensive numerical simulations. This corresponds to an omniscient oracle that provides the real performance of the candidates. The boxes on the left are only a visual aid to represent the whole parameter space (candidates in the boxes are randomly drawn from the whole space). In each iteration $i$, a random set of candidates $X\_draw\_i$ is drawn from the design space. The \textbf{Surrogate Model} predicts the performance for each candidate. Based on the predictions $y\_draw\_i$, the \textbf{Selector} chooses the best candidates $X\_aq\_i$ which are then used to improve the Surrogate Model. The expensive numerical simulation is thus only executed on already promising candidates. The cycle repeats until an abortion criterion is met, for instance, until a \textit{good-enough} candidate is found or the budget is exhausted.}
    \label{fig:experiment}
\end{figure}

We present the methodology in Figure~\ref{fig:experiment}. The DAL process starts by randomly selecting $initial\_size$ designs candidates for training $X\_train\_0$ (depicted as a grey box) from a data-pool (depicted as a blue box). Based on the design candidates $X\_train\_\{i\}$, the \textbf{Expert Model} determines the corresponding target values $y\_train\_\{i\}$. Where $i$ is the iteration loop count, indicating how many times the process has looped.

Subsequently, the design candidates and the target values are used to train the \textbf{Surrogate Model} in a supervised manner. After training, the Surrogate Model predicts the target values of a $draw\_size$ large number of design candidates $X\_draw\_\{i\}$ (depicted as green box). These predictions are passed to the \textbf{Selector}. $Draw\_size$ many random design candidates $X\_draw\_\{i\}$ are bootstrapped in every iteration. Based on the query strategy, the Selector chooses a subset of design candidates $X\_aq\_\{i\}$ with the acquisition size $aq\_size$. The Expert Model determines the true target values and the iteration loop finishes by adding the newly acquired designs to the training dataset. Each training cycle starts with newly initialized weights of the Surrogate Model. This loop iterates until a defined number $n\_iter$ is achieved. 

$\textbf{Expert Model}$: The Expert Model is not directly needed in this work, since a large annotated data-pool is available. Therefore, the Expert Model can be simulated to ensure a simple and fast pool-based experimentation and evaluation. However, the introduced experimental procedure can be used in an online setting where the Surrogate Model has to generate annotations on-the-fly.

$\textbf{Surrogate Model}$: This study utilizes a multi-layer perceptron (MLP) as the Surrogate Model, with the first hidden layer consisting of 200 neurons and the second hidden layer comprising 100 neurons. A leakyReLU activation function and a dropout layer are applied to each hidden layer to enhance the model's generalization performance. The dropout rate is set to a constant value of 0.1. For the regression task, the output layer consists of one linear neuron for each of the two target values. Both the learning rate and the batch size are kept constant. The value for the learning rate is set to $0.0005$ and the batch size to $4$. An early stopping criterion in case the training error does not reduce further after $10$ epochs is performed. The hyperparameters were determined based on results of preliminary studies. The weights of the best performing epoch are reloaded to evaluate the model's performance. At each process iteration, the model is trained from scratch to avoid potential bias to data selected in earlier iterations~\cite{hu2019active}.

\subsection{query strategies}
We developed two simple but efficient query strategies for DADO named L2-Select (L2S) and L2-Reject (L2R). These strategies can be characterized as simple in the sense that they are model-agnostic and they solely necessitate a point estimate for their targets. With these query strategies, there is no need to rely on complex, computationally expensive and sensitive methods for uncertainty modeling.
First, $draw\_size$ design candidates are bootstrapped from the entirety of the non-annotated data-pool to prevent test data leakage and to ensure an unbiased test of the model after each iteration. Subsequently, the target values $\vec{y_n}$ of these design candidates are determined by the Surrogate Model.
The goal of the strategies is to choose $aq\_size$ candidates from the target value set $y\_draw\_i$:
\begin{equation}
    y\_draw\_i = \{ \vec{y_n} | \vec{y_n} \in J_1 \times J_2\} \label{eq:ydrawi}
\end{equation}
with $|y\_draw\_i| = draw\_size$ where $J_1$ and $J_2$ represent the objectives (for the use case $J_1$: pressure loss, $J_2$: cooling performance). 

\begin{figure}
\centering
\begin{subfigure}[t]{0.24\textwidth}
    \centering
   \includegraphics[height=1.7in]{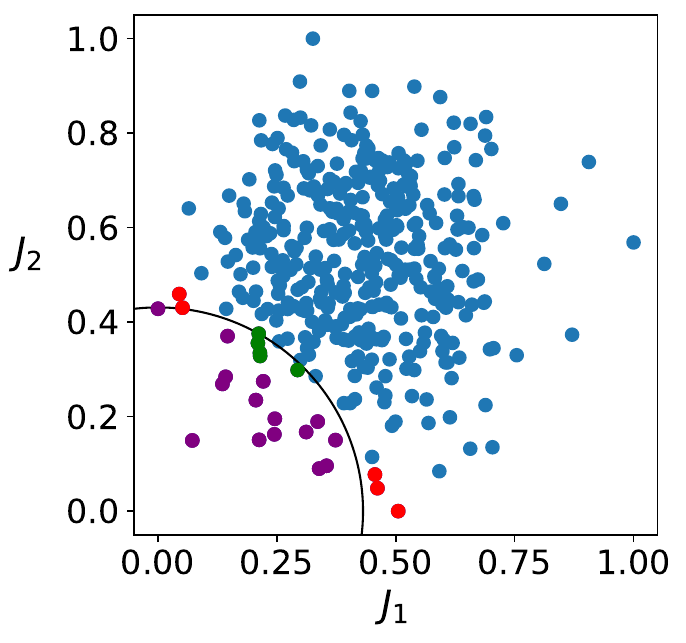}
   \caption{L2-Select}
   \label{fig:lowcost} 
\end{subfigure}
\hfill
\begin{subfigure}[t]{0.24\textwidth}
    \centering
   \includegraphics[height=1.7in]{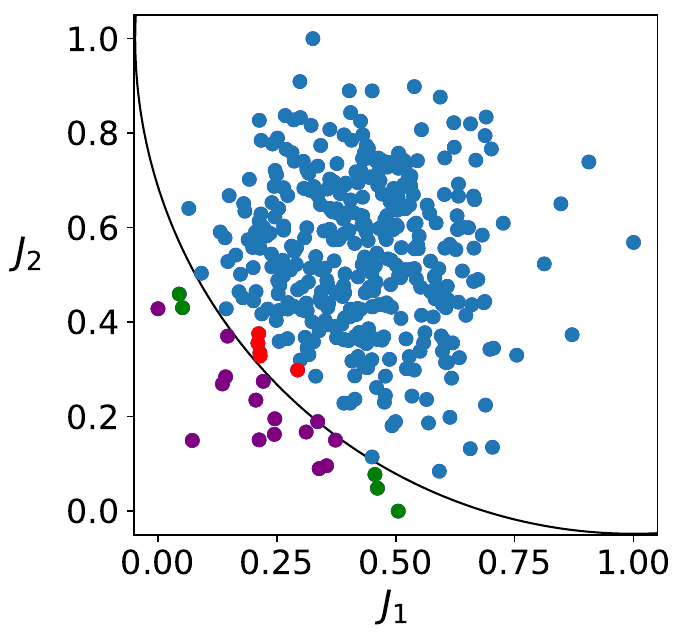}
   \caption{L2-Reject}
   \label{fig:lowcostreverse}
\end{subfigure}
    \caption{Two proposed regression query strategies for multi-objective deep active design optimization shown using a multivariate Gaussian distribution.}
    \label{fig:selection_strategies}
\end{figure}

For the L2S query strategy, the $aq\_size$ design candidates with the smallest magnitude (or L2-norm) of the target value vector~$\vec{y_n}$ are selected, cf. Equation \eqref{eq:l2s_set}. The L2-norm 
\begin{align}
L2S(\vec{y}) = |\vec{y}| = \sqrt{\sum_{j=1}^{num\_obj} y_{j}^2} \label{low_cost}\\
\{ L2S(\vec{y_n}) | \vec{y_n} \in y\_draw\_i, L2S(\vec{y_n}) \leq L2S(\vec{y_{n+1}}) \} \label{eq:l2s_set}
\end{align}
is calculated by the square root of the sum of the squared elements $y_{n,j}$ of the target vector as shown in Equation~\eqref{low_cost}. A graphical interpretation of this strategy is provided in Figure~\ref{fig:lowcost}.

L2R uses an adapted variant of the L2-norm based on the design candidate in the $draw\_size$ that has the largest predicted target values $y_{\text{max},j}$ as its origin. 
L2R uses an adapted variant of the L2-norm where the origin corresponds to the maximum values of the currently considered design candidates from $y\_draw\_i$, cf. Equation~\eqref{eq:ymax}.
Equation~\ref{low_cost_reverse} shows the adapted L2-norm and its graphical interpretation is given in Figure~\ref{fig:lowcostreverse}.
Instead of selecting the design candidate with lowest values for $L2R(\vec{y_n})$, the first $draw\_size - aq\_size$ design candidates are rejected, cf. Equation~\eqref{eq:l2r_set}. Therefore, we select the remaining $aq\_size$ design candidates which are not rejected.
\begin{align}
y_{\text{max},j} = \max\{y_{n,j}\}, \vec{y_n} \in y\_draw\_i \label{eq:ymax} \\ 
L2R(\vec{y}) = \sqrt{\sum_{j=1}^{num\_obj} (y_{j} - y_{\text{max},j})^2} \label{low_cost_reverse} \\
\{ L2R(\vec{y_n}) | \vec{y_n} \in y\_draw\_i, L2R(\vec{y_n}) \leq L2R(\vec{y_{n+1}}) \} \label{eq:l2r_set}
\end{align}
When comparing the two query strategies in more detail, the differences in the choice of design candidates can be highlighted more clearly. In Figure~\ref{fig:selection_strategies}, $400$ design candidates are plotted following a multivariate Gaussian distribution. The query strategy separates the $aq\_size$ selected design candidates from the unselected design candidates which are shown in blue. Design candidates selected by both query strategies are indicated in purple, and design candidates marked with a red or green show the differences of the selected design candidates. 

We propose the L2R as an alternative to the L2S strategy because it may offer several advantages. Firstly, we assume that L2R effectively accounts for design candidates that reside at the edges of the target space, which are often overlooked by the L2S strategy. Additionally, the selected design candidates using L2R are more likely to correspond to a Pareto front, which is a key objective in multi-objective optimization. In contrast, design candidates drawn from the core of the distribution are less likely to offer diversity in the design space, assuming L2R to be the preferred query strategy. We compare these query strategies against each other and a random selection.

\section{Experiments}
\label{sec:experimental}

\subsection{Setup}
We define two experimental scenarios. One is a \textbf{low-budget} (\textbf{S1}) experiment and the other is a \textbf{high-budget} (\textbf{S2}) one. The main difference between the two scenarios is the amount of initial design candidates $X\_train\_0$ for training and the number of design candidates $X\_aq\_i$ added to the training dataset per iteration.
\textbf{S1} holds an $initial\_size$ of $100$ design candidates, its $draw\_size$ is set to $400$ and the selected subset $aq\_size$ is variable in a set of $\{10, 20, 25, 50\}$ design candidates per iteration until a budget of $500$ design candidates is exhausted. We selected the experimental parameters for our DAL experiments based on the observation that datasets in the domain of DO are generally very small. Thus, the parameters for experiment \textbf{S1} were chosen to represent a real-world scenario.
Scenario \textbf{S2} consists of $500$ initial design candidates $X\_train\_0$ and acquires $\{50, 100, 125, 200\}$ $X\_aq\_i$ per loop execution from its $draw\_size$ of $2000$ design candidates until a budget of $1500$ is reached. In \textbf{S2}, we show the process again with a larger budget as it is typically used for DO, but the amount of data can still be considered to be very small for deep learning applications.
All experiments for the multi-objective optimization are performed using the L2S, the L2R and a random query strategy. In addition, each experiment is performed with $5$ different random seeds to ensure a representative evaluation. The different DAL experiment parameters that are investigated are summarized in Table~\ref{hyperparameter_table}.

\definecolor{LightGray}{gray}{0.9}
\definecolor{MidGray}{gray}{0.75}
\definecolor{DarkGray}{gray}{0.6}
\definecolor{Hellgrün}{RGB}{74, 172, 150}
\definecolor{Hellblau}{RGB}{80, 149, 200}

\begin{table*}[]
\caption{The parameters and their values used in the experiments for both scenarios.}
\begin{center}
\begin{tabular}{c|c|c|c|c|c|c|c|c}
\textbf{Scenarios}& \multicolumn{4}{c|}{\cellcolor{Hellgrün} \textbf{Low\_Budget~S1}}&\multicolumn{4}{c}{\cellcolor{Hellblau} \textbf{High\_Budget~S2}}\\
\cline{1-9} 
\textbf{\textit{query\_strategy}} & random & L2-Select &\multicolumn{2}{c|}{ L2-Reject} & random & L2-Select &\multicolumn{2}{c}{ L2-Reject} \\
\cline{1-9} 
\textbf{\textit{initial\_size}}&\multicolumn{4}{c|}{$100$}&\multicolumn{4}{c}{$500$} \\
\cline{1-9}
\textbf{\textit{draw\_size}}&\multicolumn{4}{c|}{$400$}&\multicolumn{4}{c}{$2000$} \\
\cline{1-9}
\textbf{\textit{aq\_size}}&$10$&$20$&$25$&$50$&$50$&$100$&$125$&$200$\\
\cline{1-9} 
\textbf{\textit{budget}}&\multicolumn{4}{c|}{$500$}&\multicolumn{4}{c}{$1500$} \\
\end{tabular}
\label{hyperparameter_table}
\end{center}
\end{table*}

To evaluate the experiments presented, we employ various metrics, including the mean square error (MSE), the spearman rank-order correlation coefficient (SROCC), as well as the mean rank (MR) and intersection metrics, which we introduce below. The results of the conducted experiments are summarized with the help of the area under the learning curve (AUC) metric in Table~\ref{tab:results_table}. This allows us to evaluate an entire optimization process in a single value per metric. To calculate the SROCC and MR metrics, it is necessary to sort the target value $\vec{y_n}$ of the estimated $y\_draw\_i$ and the true values $y\_draw\_i_{true}$ according to their quantity. 
To do so, we sort the currently drawn candidates $X\_draw\_i$ based in the current query strategy $S$ (i.e., L2S, L2R, random) and the true performance values, cf. Equation~\eqref{eq:mrset}.
The set $K$ contains the indices from the sorted candidates set that correspond to the candidates in $X\_aq$.
The MR metric is then the average of the first $aq\_size$ indices of the set $K$.
Additionally, we normalized the MR to be between 0 and 1, where 0 corresponds to its optimal value depending on its $aq\_size$ and the MR after the first process iteration which is to be assumed the highest value of the process. 
The optimal value of the MR metric would therefore result in $aq\_size \cdot 1/2$,
\begin{align}
    K =& \{ k | \vec{x_n} = \vec{z_k}, \vec{x_n} \in X\_aq, \notag \\
    &\vec{z_k} \in \text{sort}(X\_draw\_i, S(y\_draw\_i_{true})) \} \label{eq:mrset}\\
    \textbf{MR} =& \frac{1}{aq\_size} \sum_{k \in K} k \label{eq:mr}
\end{align}
The SROCC is calculated using the first $aq\_size$ indices of both sorted lists as input and outputs a value between~$0$~and~$1$, where a value of $1$ indicates that the $aq\_size$ design candidates of the predicted values match the correct sorting of the true values. 
The intersections metric assesses the accuracy of the top-rated designs. 
The metric is relatively simple. It compares the $aq\_size$ selected candidates $X\_aq\_i$ 
against $ X\_aq\_i_{true} \subset X\_draw\_i$, the $aq\_size$ candidate selected based on the ground truth performance.
The intersection of both sets can be used to directly calculate the accuracy which is based on the cardinality of the intersection, cf. Equation~\eqref{eq:inter}.
The name \textit{intersection} for the metric is based on the intersection operation.
\begin{align}
    \textbf{intersection} = \frac{|X\_aq\_i \cap X\_aq\_i_{true}|}{aq\_size} \label{eq:inter}
\end{align}

We prioritize SROCC, MR, and intersections metrics over classic MSE for DADO, as accurate ranking of designs is more crucial than precise estimations of target values. With the true ranking of the designs, the true $y\_aq\_\{i\}$ values are calculated using the Expert Model. Nevertheless, we assume that DAL will lead to an improvement of the MSE of the added designs $X\_aq\_\{i\}$ after each iteration.

\subsection{Results}
In Table \ref{tab:results_table}, we present the AUC and the final value at the end of the process for all experiments and metrics. \textbf{S1} is highlighted in green and \textbf{S2} in blue, while the three query strategies are differentiated by varying shades of gray. The results indicate that L2S outperforms L2R in every single experiment and that the random strategy consistently yields the worst results, except in the case of rnd\_MSE which is the MSE between $y\_draw\_i$ and its true annotations. Additionally, the quality of the results does not simply increase with its $aq\_size$, for \textbf{S1}. For \textbf{S2}, the experiment with the smallest $aq\_size$ based on the best\_MSE, the MR and the SROCC provide predictions with the highest performance. The best\_MSE is the MSE between $y\_aq\_i$ and its true annotations.

\begin{table*}[]
\caption{The results of the conducted experiments are summarized. The final value of each metric corresponds to the value obtained at the end of the DAL cycle. In addition, the overall performance of the cycle has been quantified using the AUC, which captures the balance between the number of queries and the improvement in the model's performance over the course of the cycle.}
\begin{center}
\begin{tabular}{c|c|c|c|c|c|c|c|c|c|c|c}
\textbf{AQ\_Size} & \textbf{query\_strategy }&  \multicolumn{2}{c|}{\textbf{intersections}} & \multicolumn{2}{c|}{\textbf{ mean\_rank }}& \multicolumn{2}{c|}{\textbf{ srocc\_rank }}& \multicolumn{2}{c|}{\textbf{ best\_MSE}} &  \multicolumn{2}{c}{\textbf{ rnd\_MSE}}\\
\cline{1-12}
                                                 &                    &          AUC &                Final &          AUC &                Final &          AUC &                Final &          AUC &                Final &          AUC &                Final  \\ 
\cline{1-12}
\rowcolor{LightGray} \cellcolor{Hellgrün} & random & 0.022 & 0.020 & 0.268 & 0.117 & 0.201 & 0.319 & 0.093 & 0.024 & 0.083 & 0.066 \\         
\rowcolor{MidGray} \cellcolor{Hellgrün} & L2-Reject & 0.316 & 0.460 & 0.327 & 0.149 & 0.309 & 0.605 & 0.021 & 0.004 & 0.108 & 0.088 \\          
\rowcolor{DarkGray} \multirow{-3}{*}{{\cellcolor{Hellgrün}10}} & L2-Select & 0.394 & 0.560 & 0.292 & 0.064 & 0.353 & 0.658 & 0.022 & 0.002 & 0.106 & 0.094 \\      
\hline
\rowcolor{LightGray}\cellcolor{Hellgrün}        &             random &        0.044 &                0.050 &      0.337 &            0.138 &       0.262 &             0.275 &     0.110 &           0.031 &    0.083 &          0.055 \\      
\rowcolor{MidGray}\cellcolor{Hellgrün}        &  L2-Reject &        0.380 &                0.460 &      0.309 &            0.151 &       0.338 &             0.429 &     0.032 &           0.009 &    0.102 &          0.084 \\          
\rowcolor{DarkGray} \multirow{-3}{*}{{\cellcolor{Hellgrün}20}}       &          L2-Select &        0.445 &                0.570 &      0.303 &            0.089 &       0.335 &             0.444 &     0.031 &           0.005 &    0.096 &          0.106 \\  \cline{1-12}        
\rowcolor{LightGray}\cellcolor{Hellgrün}        &             random &        0.061 &                0.096 &      0.350 &            0.130 &       0.282 &             0.384 &     0.153 &           0.018 &    0.086 &          0.049 \\          
\rowcolor{MidGray}\cellcolor{Hellgrün}        &  L2-Reject &        0.417 &                0.504 &      0.325 &            0.152 &       0.312 &             0.556 &     0.069 &           0.003 &    0.100 &          0.083 \\          
\rowcolor{DarkGray} \multirow{-3}{*}{{\cellcolor{Hellgrün}25}} &          L2-Select &        0.481 &                0.592 &      0.371 &            0.155 &       0.391 &             0.668 &     0.029 &           0.004 &    0.103 &          0.082 \\ 
\hline
\rowcolor{LightGray}\cellcolor{Hellgrün}       &             random &        0.119 &                0.124 &      0.415 &            0.216 &       0.314 &             0.510 &     0.138 &           0.224 &    0.085 &          0.077 \\          
\rowcolor{MidGray}\cellcolor{Hellgrün}       &  L2-Reject &        0.483 &                0.604 &      0.403 &            0.250 &       0.370 &             0.574 &     0.039 &           0.006 &    0.098 &          0.072 \\          
\rowcolor{DarkGray} \multirow{-3}{*}{{\cellcolor{Hellgrün}50}} &          L2-Select &        0.546 &                0.676 &      0.423 &            0.195 &       0.372 &             0.557 &     0.040 &           0.008 &    0.107 &          0.082 \\
\hline
\hline
\rowcolor{LightGray}\cellcolor{Hellblau}        &               random &      0.023 &                0.024 &      0.610 &            0.611 &       0.395 &             0.423 &     0.072 &           0.063 &    0.040 &          0.025 \\          
\rowcolor{MidGray}\cellcolor{Hellblau}        &    L2-Reject &      0.509 &                0.576 &      0.587 &            0.301 &       0.603 &             0.737 &     0.001 &           0.000 &    0.069 &          0.065 \\          
\rowcolor{DarkGray} \multirow{-3}{*}{{\cellcolor{Hellblau}50}}       &            L2-Select &      0.699 &                0.824 &      0.396 &            0.137 &       0.686 &             0.859 &     0.002 &           0.001 &    0.067 &          0.070 \\   
\hline
\rowcolor{LightGray}\cellcolor{Hellblau}       &               random &      0.046 &                0.050 &      0.716 &            0.512 &       0.478 &             0.542 &     0.064 &           0.026 &    0.042 &          0.024 \\          
\rowcolor{MidGray}\cellcolor{Hellblau}       &    L2-Reject &      0.556 &                0.664 &      0.575 &            0.304 &       0.617 &             0.720 &     0.002 &           0.001 &    0.069 &          0.056 \\          
\rowcolor{DarkGray} \multirow{-3}{*}{{\cellcolor{Hellblau}100}}        &            L2-Select &      0.714 &                0.780 &      0.516 &            0.246 &       0.677 &             0.806 &     0.003 &           0.001 &    0.067 &          0.057 \\   
\hline
\rowcolor{LightGray}\cellcolor{Hellblau}       &               random &      0.060 &                0.066 &      0.712 &            0.564 &       0.505 &             0.513 &     0.066 &           0.017 &    0.043 &          0.027 \\          
\rowcolor{MidGray}\cellcolor{Hellblau}       &    L2-Reject &      0.574 &                0.654 &      0.647 &            0.425 &       0.639 &             0.778 &     0.002 &           0.001 &    0.068 &          0.060 \\          
\rowcolor{DarkGray} \multirow{-3}{*}{{\cellcolor{Hellblau}125}} &            L2-Select &      0.721 &                0.802 &      0.462 &            0.215 &       0.679 &             0.814 &     0.003 &           0.001 &    0.071 &          0.058 \\     
\hline
\rowcolor{LightGray}\cellcolor{Hellblau}       &               random &      0.104 &                0.107 &      0.675 &            0.469 &       0.570 &             0.602 &     0.051 &           0.027 &    0.045 &          0.030 \\          
\rowcolor{MidGray}\cellcolor{Hellblau}       &    L2-Reject &      0.610 &                0.653 &      0.630 &            0.486 &       0.624 &             0.703 &     0.003 &           0.002 &    0.065 &          0.067 \\          
\rowcolor{DarkGray} \multirow{-3}{*}{{\cellcolor{Hellblau}200}}       &            L2-Select &      0.736 &                0.802 &      0.556 &            0.282 &       0.650 &             0.751 &     0.005 &           0.002 &    0.068 &          0.061 \\          
\end{tabular}
\label{tab:results_table}
\end{center}
\end{table*}

\begin{figure}[]
\centering
\begin{subfigure}[t]{0.24\textwidth}
    \centering
  \includegraphics[height=1.7in]{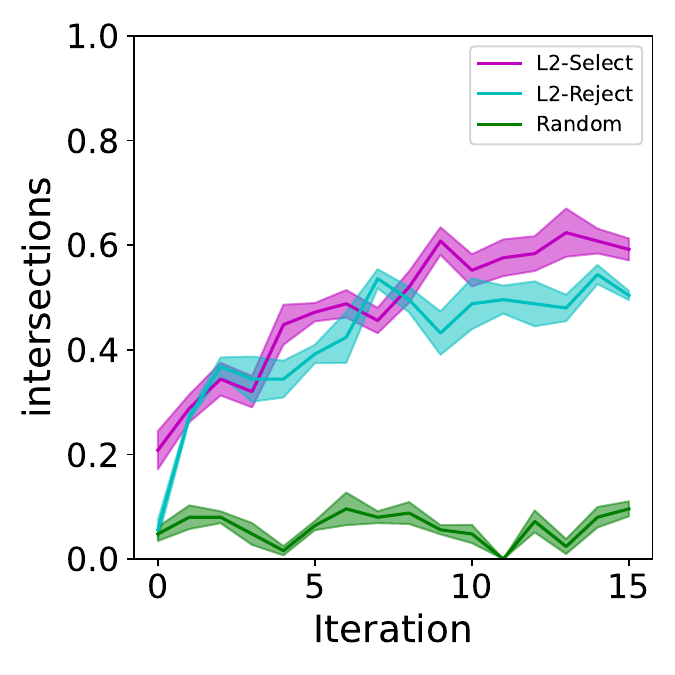}
   \caption{Intersections}
   \label{fig:IntersectionsS1} 
\end{subfigure}
\hfill
\begin{subfigure}[t]{0.24\textwidth}
    \centering
    \includegraphics[height=1.7in]{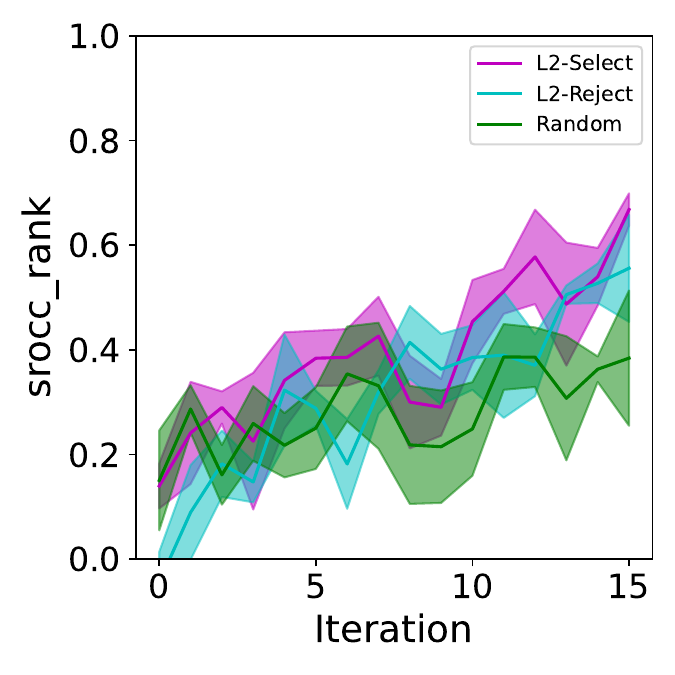}
   \caption{SROCC}
   \label{fig:SROCCS1}
\end{subfigure}
\begin{subfigure}[t]{0.24\textwidth}
    \centering
    \includegraphics[height=1.7in]{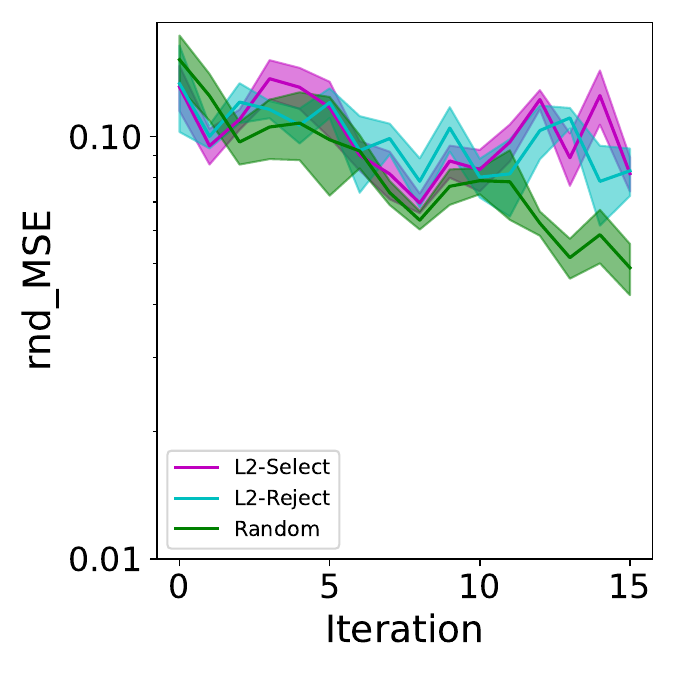}
   \caption{rnd\_MSE}
   \label{fig:rndMSES1} 
\end{subfigure}
\hfill
\begin{subfigure}[t]{0.24\textwidth}
    \centering
    \includegraphics[height=1.7in]{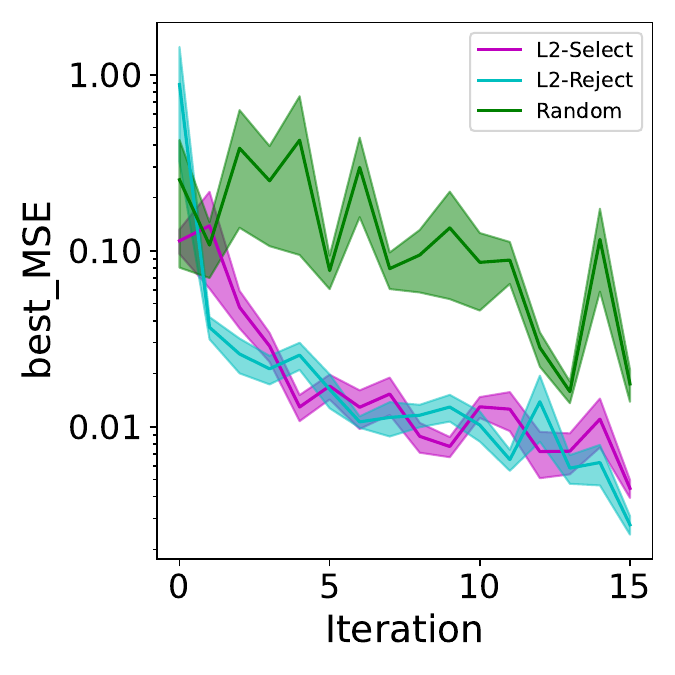}
   \caption{best\_MSE}
   \label{fig:bestMSES1}
\end{subfigure}
    \caption{Evaluation of the low-budget experiment \textbf{S1} with $aq\_size$ of 25. The rnd\_MSE represents the MSE between $y\_draw\_i$ and its true annotations, while best\_MSE denotes the MSE between $y\_aq\_i$ and its true annotations.}
    \label{fig:lowBudget}
\end{figure}

\begin{figure}[]
\centering
\begin{subfigure}[t]{0.235\textwidth}
    \centering
   \includegraphics[height=1.7in]{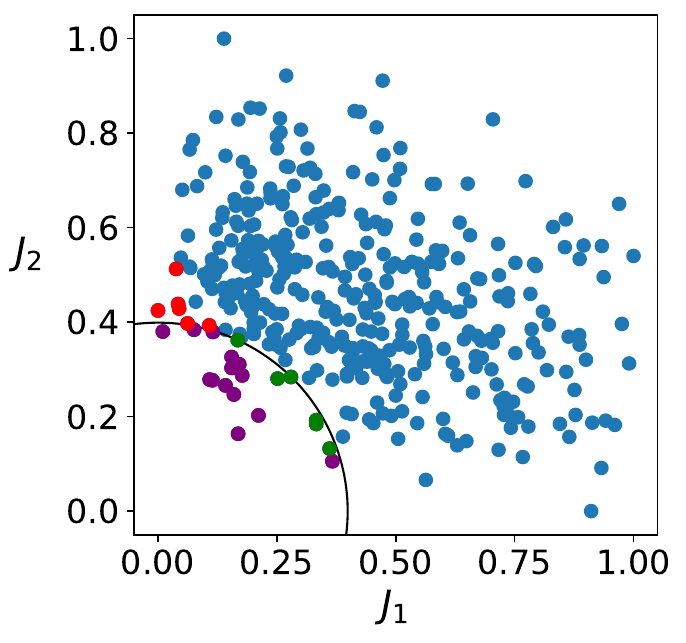}
   \caption{L2-Select}
   \label{fig:low_cost_real} 
\end{subfigure}
\hfill
\begin{subfigure}[t]{0.235\textwidth}
    \centering
   \includegraphics[height=1.7in]{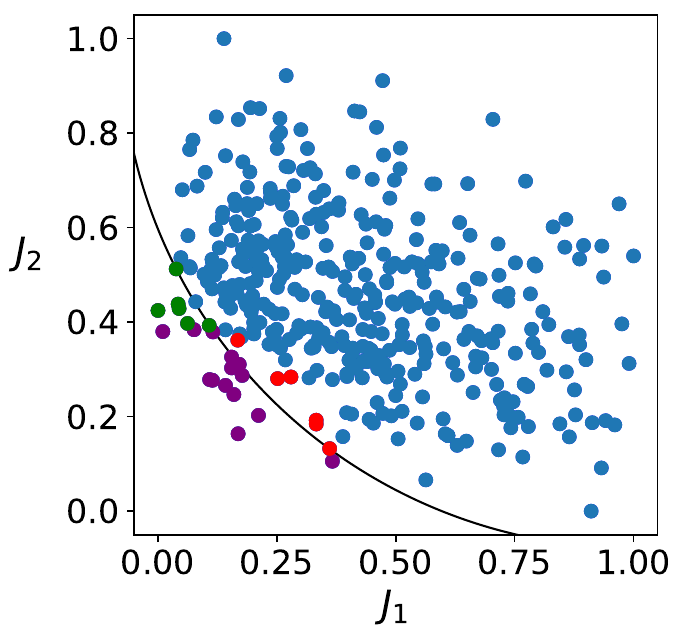}
   \caption{L2-Reject}
   \label{fig:lowcostreverse_real}
\end{subfigure}
    \caption{Two proposed regression query strategies for multi-objective deep active design optimization. Shown on 200 randomly drawn design candidates from our data-pool.}
    \label{fig:selection_strategies_real}
\end{figure}

\begin{figure}[H]
\centering
\begin{subfigure}[t]{0.24\textwidth}
    \centering
   \includegraphics[height=1.7in]{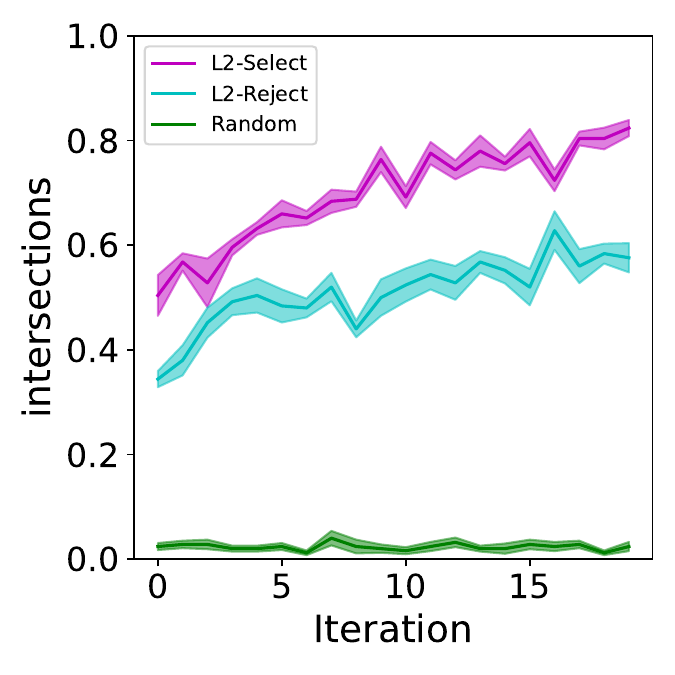}
   \caption{Intersections}
   \label{fig:IntersectionsS2} 
\end{subfigure}
\hfill
\begin{subfigure}[t]{0.24\textwidth}
    \centering
      \includegraphics[height=1.7in]{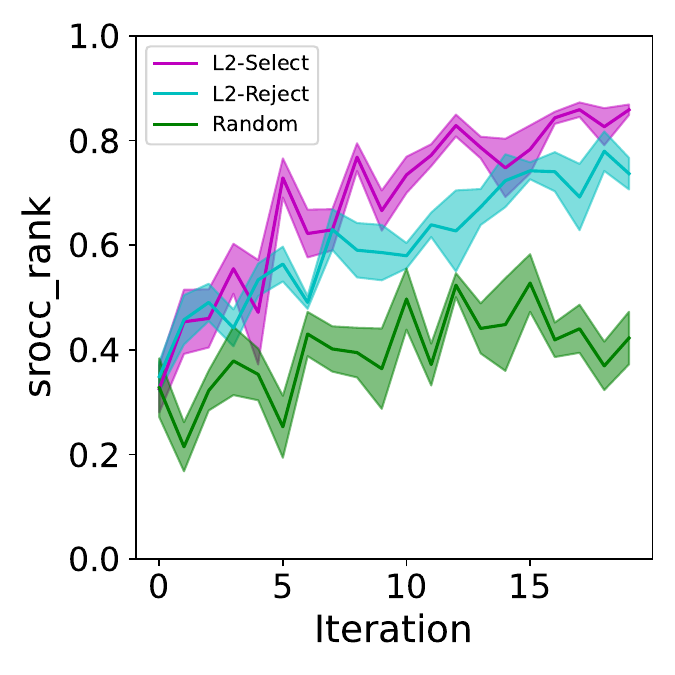}
   \caption{SROCC}
   \label{fig:SROCCS2}
\end{subfigure}
\begin{subfigure}[t]{0.24\textwidth}
    \centering
         \includegraphics[height=1.7in]{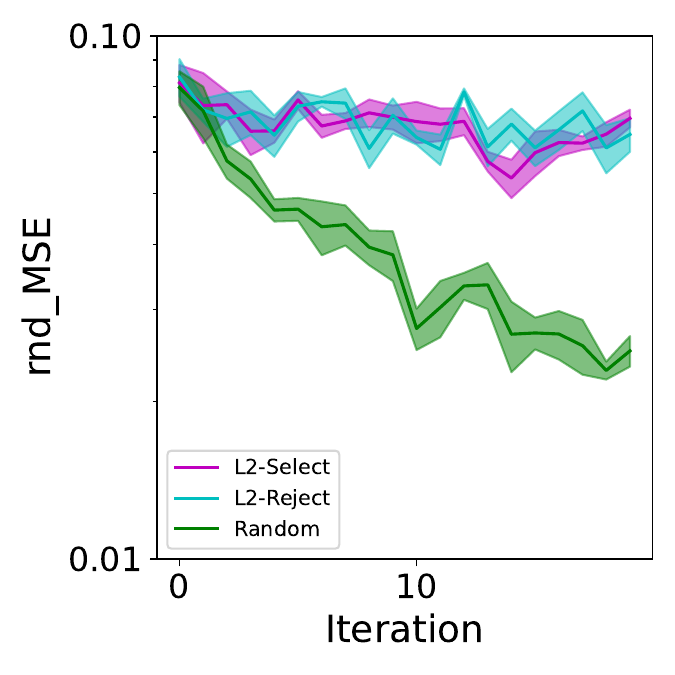}
   \caption{rnd\_MSE }
   \label{fig:rndMSES2} 
\end{subfigure}
\hfill
\begin{subfigure}[t]{0.24\textwidth}
    \centering
    \includegraphics[height=1.7in]{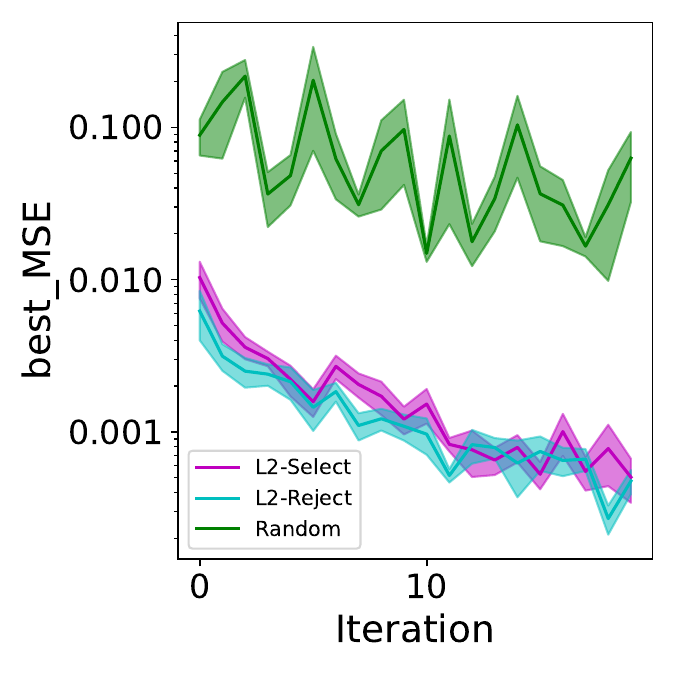}
   \caption{best\_MSE}
   \label{fig:bestMSES2}
\end{subfigure}
    \caption{Evaluation of the high-budget experiment \textbf{S2} with $aq\_size$ of 50. The rnd\_MSE represents the MSE between $y\_draw\_i$ and its true annotations, while best\_MSE denotes the MSE between $y\_aq\_i$ and its true annotations.}
    \label{fig:highBudget}
\end{figure}

The superiority of the random query strategy in the rnd\_MSE metric is attributable to the bias that we attempt to impose through our query strategies, whereby the model's predictions in regions where the query strategies assume promising values are expected to yield a higher performance. As such, it is reasonable that models that are trained using the design candidates suggested by L2S and L2R would perform worse in other regions. The best\_MSE identifies the MSE that was evaluated on the selected design candidates. This metric monitors the predictive performance of our model. To look at the results in more detail, we have selected an experiment for \textbf{S1} and \textbf{S2} which we would like to discuss in more detail below. In Figure~\ref{fig:lowBudget} shows the results for \textbf{S1} with a $aq\_size$ of 25 design candidates per iteration. Since our $initial\_size$ is 100 design candidates and our budget is 500 design candidates in total, we iterate 16 times.
In Figure~\ref{fig:IntersectionsS1}, we present the intersections metric, the SROCC in Figure~\ref{fig:SROCCS1}, the rnd\_MSE in Figure~\ref{fig:rndMSES1}, and the best\_MSE in Figure~\ref{fig:bestMSES1}. The lines represent the mean values obtained from the five runs conducted for each experiment. In addition to the mean values, the plot displays the standard-error intervals for each metric and query strategy. Throughout the course of the experiment, it is evident that the intersections and the SROCC show an increasing trend, while the rnd\_MSE and the best\_MSE exhibit a decreasing trend. Although the random strategy shows good predictive performance on the randomly selected design candidates, it underperforms compared to the two query strategies on the promising design candidates. The benefits of the low-cost query strategies become apparent upon examining the following metrics. The intersection metric shows that the process develops a self-awareness in the course of the iterations and is increasingly able to select suitable design candidates for multi-objective DO. However, it becomes apparent that the intersection metric is too strict for random selection, and despite being able to improve, models trained on the basis of random selection are still unable to satisfy this metric. Therefore, the MR and the SROCC are introduced as alternative metrics. While the visualization of the MR has been omitted due to limited space, it is proven to be a useful metric for comparing experiments (see Table~\ref{tab:results_table}). The SROCC shows a similar qualitative trend as the intersection metric, with L2S outperforming L2R and the random strategy. However, it also reveals that the random strategy improves in sorting the $draw\_size$ design candidates by rank based on their target value over the iterations, which is not reflected by the intersection metric. Unexpectedly, the L2S strategy outperforms the L2R strategy, which may be attributed to the nature of the available data. The query strategy was originally designed for a multivariate Gaussian distribution; however, as illustrated in Figure~\ref{fig:selection_strategies_real}, the two scaled target values of the real data do not conform to a Gaussian distribution, hence the solution quality of L2S exceeds that of L2R in this use case.
As stated before, the \textbf{S2} experiment with an $aq\_size$ of 50 produced the best results. Therefore, we will examine the results of this experiment more closely in the Figure~\ref{fig:highBudget}. Since \textbf{S2} had a budget of 1500 design candidates, this implies that 20 iterations were completed.

When examining the results from \textbf{S2}, it becomes evident that the standard-error intervals in the experiments are considerably reduced due to the larger budget. Additionally, the metrics are notably improved when compared to \textbf{S1}. The disparities between the query strategies mentioned earlier are also more distinct in evaluating \textbf{S2} but are in line with the outcomes previously discussed for \textbf{S1}. Detecting any noticeable variation in prediction quality based on the rnd\_MSE is challenging for both L2S and L2R. The best\_MSE values exhibit almost identical patterns and trends. Nonetheless, differences in the performance between L2R and L2S can be observed with the aid of the intersection and the SROCC metrics. Notably, a decline in the slope of the curve can be inferred with random selection, as indicated by the SROCC. Also noteworthy is the high fluctuation of the best\_MSE in random selection, from which it can be concluded that the prediction performance on the selected design candidates is considerably lower.  

In comparing the four metrics between the final iteration of \textbf{S1} (cf. Figure~\ref{fig:lowBudget}) and the initial iteration of \textbf{S2} (cf. Figure~\ref{fig:highBudget}), a considerable performance improvement is observed in favor of \textbf{S1}, despite both scenarios having an equal budget at that stage. This finding supports the effectiveness and benefits of our methodology in DO

\section{Towards Generative Deep Active Design Optimization}
\label{sec:outlook}
Based on our confidence in the feasibility of performing self-optimizing multi-objective optimization using DAL, we aim to augment the Surrogate Model in the presented process with a VAE. Similar to Parekh et al.~\cite{parekh2022Variational}, we extend the VAE with an additional prediction network and, thereby, perform a multi-task regression and reconstruction model. As described in Section~\ref{sec:related_work}, we believe that their VAE is exclusive learning the identity of the two motor topologies. The reason for that is the chosen size of the latent space and the fact that the used trainingset does not represent a real-world DO scenario.

Our idea is to embed the VAE into the DAL process presented above. As a query strategy, a clustering approach in the latent representation shall be applied to separate areas of promising design candidates from other less well-performing design candidates. The generative properties of the VAE will then be used to specifically generate new design candidates which belong to the promising area of latent space. The smaller the latent size is, the easier it will be for clustering methods to separate these areas, but the more challenging the subsequent reconstruction of the design candidates might be. The prediction network is parallel to the decoder of the VAE in the latent space. With the help of this additional network, the latent space can be divided based on the predicted target values, to enable clustering.

\begin{figure}[ht!]
    \centering
    \begin{tikzpicture}
        \begin{axis}[
            paraviewStyle, 
            colorbar horizontal,
            colormap/blackwhite, 
            colorbar style={
            title={Absolute error [-]}, 
            width=0.4\textwidth, 
            height=6, 
            xtick={0,0.1,0.2,0.3,0.4,0.5},
            point meta min=-0, 
            point meta max=0.55, 
            }
            ]
        \end{axis}
    \end{tikzpicture}
    \includegraphics[width=.48\textwidth]{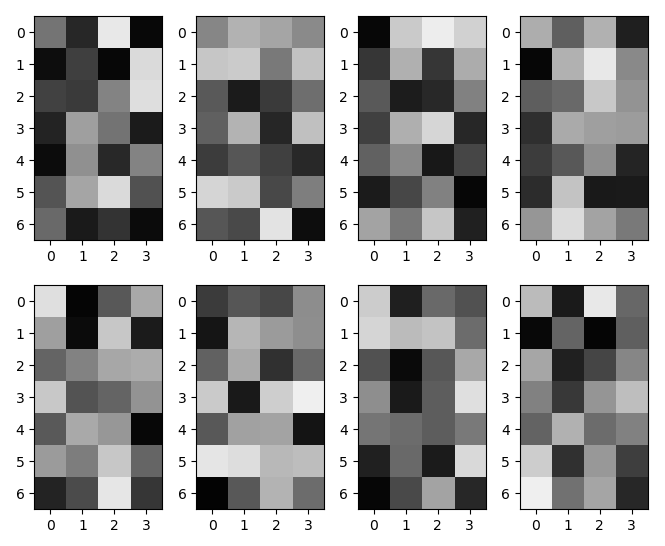}
    \caption{Reconstructed test design candidates selected at random using the VAE model. The 28 design parameters are presented in a 4x7 grid configuration, enabling visualization of the absolute deviation between the reconstructed values and the ground truth targets. The color gradient indicates the magnitude of the deviation, with darker shades representing superior reconstruction performance.}
    \label{fig:VAE_samples}
\end{figure}
 
Although numerous experiments have been conducted to optimize the structure and hyperparameters of the VAE, a suitable trade-off between a well-separable latent space and a reconstruction error, that is not excessively large, has yet to be found. If the reconstruction error is too large, it is expected that there will be a high deviation between the prediction of the Surrogate Model and the Expert Model. On the other hand, if the prediction performance is too low, patterns cannot be detected in the latent space. This issue may be attributed to the weighting factor $\beta$, which determines the influence of the Kullback Leibler divergence. Several approaches have been explored, such as introducing a cyclical annealing schedule~\cite{fu2019cyclical} for $\beta$, to address the reconstruction and separability trade-off. However, no clear trend can be observed in Figure~\ref{fig:VAE_samples}, which displays the absolute deviation of the reconstruction of eight random test design candidates. Each of the 28 boxes, arranged in a 7x4 grid, represents a design parameter.

In our next steps, we plan to introduce cyclic training for the reconstruction and prediction tasks. We believe that the VAE, as a Surrogate Model, will play a central role in DADO, and therefore, we consider it the focus of our future work.

\section{Conclusion}
\label{sec:conclusion}
In this experience report, we have demonstrated the feasibility of utilizing DAL to tackle DO, leveraging a large pool of non-annotated data. The developed DAL query strategies for regression applied to a multi-objective DO have shown promising results. The results remain consistent across all of the experiments. The conducted experiments show that based on rnd\_MSE metric the performance of the random query strategy surpasses both of our query strategies. This outcome is, however, unsurprising, as the MSE of the prediction is computed based on randomly selected design candidates from the entirety of our data-pool. Nevertheless, our objective is to bias the model to perform well on the self-selected design candidates, which are advantageous for self-optimization by proposing promising design candidates only. Our assumption that the developed L2R query strategy would outperform the L2S strategy due to its more Pareto-like selection was not confirmed by the experiments. The reason for this is that the method was developed using a multivariate Gaussian distribution. In our dataset, the assumption of a Gaussian is not fulfilled. Especially, for small $draw\_sizes$. Both strategies presented in this article rely on the L2-norm, which selects a sample circularly around an origin. To improve the robustness of the selection to differently scaled target values, one possibility is to replace the circular selection with an ellipsoid. Our study demonstrated that the query strategies are providing promising results with two target values, an extension to higher dimensional multi-objective optimization should be straightforward. 

Further, we have shown the limitations of incorporating a generative model into the DADO process. We plan to develop a query strategy based on a clustering procedure in the latent space, once we have achieved a good balance between reconstruction and disentanglement in the latent space.\\
Subsequently, we propose the integration of the complex numerical simulations into the process, described in this work, enabling real-time generation of annotations for design candidates outside the existing data-pool. Moreover, there is clear potential in exploring the Surrogate Model and its hyperparameters to enhance prediction quality and accelerate DO, which was not the focus of this study. An investigation of the raw data that is available in numerical simulations in the sense of numerical meshes could be investigated using graph neural networks in order to determine if another data representation is advantageous for performance predictions in DADO.

With our research endeavors, we seek to make a contribution towards the reduction of CFD  and FEA simulations on HPC systems. By reducing the number of such simulations, we aim to effectively reduce the associated energy costs and mitigate associated climate-damaging emissions, thus promoting a more sustainable and environmentally conscious approach for future computational simulations.

\section*{Acknowledgment}
We express our gratitude to Dr.~Franz~Götz-Hahn for the insightful discussions. 
\bibliographystyle{ieeetr}
\bibliography{refs}

\end{document}